\def\year{2022}\relax
\documentclass[letterpaper]{article} % DO NOT CHANGE THIS

\usepackage{amssymb, amsmath, amsthm}
\usepackage[polish,english]{babel}
\usepackage[T1,nomathsymbols]{polski}
\usepackage[utf8]{inputenc}
\usepackage{multirow}
\usepackage{aaai22}  % DO NOT CHANGE THIS
\usepackage{times}  % DO NOT CHANGE THIS
\usepackage{helvet} % DO NOT CHANGE THIS
\usepackage{courier}  % DO NOT CHANGE THIS
\usepackage[hyphens]{url}  % DO NOT CHANGE THIS
\usepackage{graphicx} % DO NOT CHANGE THIS
\usepackage{natbib}  % DO NOT CHANGE THIS AND DO NOT ADD ANY OPTIONS TO IT
\usepackage{caption} % DO NOT CHANGE THIS AND DO NOT ADD ANY OPTIONS TO IT
\title{MeTeoR: Practical Reasoning in Datalog with Metric Temporal Operators}
\author{Dingmin Wang\textsuperscript{\rm 1}, Pan Hu\textsuperscript{\rm 1,2}, Przemysław Andrzej Wałęga\textsuperscript{\rm 1}, Bernardo Cuenca Grau\textsuperscript{\rm 1}
}
\affiliations{
\textsuperscript{\rm 1}Department of Computer Science, University of Oxford, UK\\
\{dingmin.wang, przemyslaw.walega, bernardo.cuenca.grau\}@{cs.ox.ac.uk} \\

\textsuperscript{\rm 2}School of Electronic Information and Electrical Engineering, Shanghai Jiao Tong University, China\\
pan.hu@{sjtu.edu.cn}
}

\usepackage{amssymb, amsmath}

\usepackage[polish,english]{babel}
\usepackage[T1,nomathsymbols]{polski}
\usepackage[utf8]{inputenc}

\usepackage{stmaryrd}
\usepackage{mathtools}
\usepackage[geometry]{ifsym}
\usepackage{paralist}
\usepackage{makecell}
\usepackage{color, colortbl}
\usepackage{multirow}
\usepackage{multicol}
\usepackage{mathrsfs}
\usepackage{caption}
\usepackage{subcaption}

\usepackage{cleveref}
\usepackage{tikz}
\usetikzlibrary{arrows,backgrounds,automata,topaths,patterns,decorations.pathreplacing,decorations.pathmorphing}
\usetikzlibrary{arrows, chains, positioning, quotes, shapes.geometric}
\usetikzlibrary{calc}
\usetikzlibrary{fit}
\usetikzlibrary{arrows}
\usetikzlibrary{positioning,automata}

\newcommand{\PS}{\ensuremath{\textsc{\textup{PSpace}}}}
\newcommand{\EXPS}{\ensuremath{\textsc{\textup{ExpSpace}}}}

\newcommand{\MTL}{\ensuremath{\textup{DatalogMTL}}}

\newcommand{\matA}{M}

\newcommand{\sbf}{\mathbf{s}}

\DeclareFontFamily{U}{MnSymbolC}{}
\DeclareSymbolFont{MnSyC}{U}{MnSymbolC}{m}{n}

\DeclareMathSymbol{\square}{\mathbin}{MnSyC}{106}
\DeclareMathSymbol{\meddiamond}{\mathbin}{MnSyC}{110}
\DeclareMathSymbol{\boxminus}{\mathbin}{MnSyC}{112}
\DeclareMathSymbol{\boxplus}{\mathbin}{MnSyC}{116}
\DeclareMathSymbol{\diamondminus}{\mathbin}{MnSyC}{120}
\DeclareMathSymbol{\diamondplus}{\mathbin}{MnSyC}{124}

\DeclareFontShape{U}{MnSymbolC}{m}{n}{
    <-5>  MnSymbolC4
    <5-6>  MnSymbolC5
   <6-7>  MnSymbolC6
   <7-8>  MnSymbolC7
   <8-9>  MnSymbolC8
   <9-10> MnSymbolC9
  <10-12> MnSymbolC10
  <12->   MnSymbolC12}{}

\newcommand{\Si}{\mathcal{S}}
\newcommand{\Ui}{\mathcal{U}}

\newcommand{\I}{\mathfrak{I}}
\newcommand{\D}{\mathcal{D}}

\newcommand{\can}[2]{\mathfrak{C}_{#1,#2}}

\newcommand{\ground}[2]{\mathsf{ground}(#1,#2)}

\newcommand{\Prog}{\Pi}

\usepackage{todonotes}

\makeatletter
\newtheorem*{rep@theorem}{\rep@title}
\newcommand{\newreptheorem}[2]{%
	\newenvironment{rep#1}[1]{%
		\def\rep@title{#2 \ref{##1}}%
		\begin{rep@theorem}}%
		{\end{rep@theorem}}}
\makeatother

\newreptheorem{theorem}{Theorem}
\newreptheorem{proposition}{Proposition}
\newreptheorem{lemma}{Lemma}

\usepackage{graphicx}        % standard LaTeX graphics tool
\usepackage{multicol}        % used for the two-column index
\usepackage[bottom]{footmisc}% places footnotes at page bottom
\usepackage{url}
\usepackage{pifont}
\usepackage{booktabs}
\usepackage{tabularx}
\usepackage{newtxtext}       %
\usepackage{newtxmath}       % selects Times Roman as basic font
\usepackage[ruled,linesnumbered,noend]{algorithm2e}
\usepackage{makecell}
\usepackage{anyfontsize}

\usepackage{tikz}
\usepackage{pgfplots}
\usetikzlibrary{arrows,backgrounds,automata,topaths,patterns,decorations.pathreplacing,decorations.pathmorphing}
\usetikzlibrary{arrows, chains, positioning, quotes, shapes.geometric}
\usetikzlibrary{calc}
\usetikzlibrary{fit}
\usetikzlibrary{arrows}
\usetikzlibrary{positioning,automata}

\definecolor{Green}{RGB}{0,144,0}

\usepackage{amssymb, amsmath, amsthm}

\theoremstyle{definition}

\begin{document}
%	\linenumbers

\maketitle

\begin{abstract}
DatalogMTL is an extension of Datalog with operators from
metric temporal logic which has received significant attention in recent years.
It is a highly expressive knowledge representation
language that is well-suited for  applications
in temporal ontology-based query answering and stream processing.
Reasoning in DatalogMTL is, however, of  high computational complexity, making
implementation  challenging and hindering its adoption
in applications.
In this paper, we present a novel approach for practical reasoning
in DatalogMTL which combines
materialisation (a.k.a.\ forward chaining)
with automata-based techniques. We have implemented
this approach in a
reasoner called MeTeoR and evaluated its performance using a temporal extension of
the Lehigh University Benchmark and a benchmark  based on
real-world  meteorological data. Our experiments show that MeTeoR is a scalable
system which enables reasoning over
complex temporal rules and  datasets involving tens of millions of temporal facts.
\end{abstract}

\section{Introduction}

Temporal data is ubiquitous in many application scenarios, such as stock
trading~\cite{nuti2011algorithmic}, network flow anomaly
detection~\cite{munz2007real}, and equipment malfunction
monitoring~\cite{doherty2009temporal}. 
In order to represent knowledge and subsequently reason in the presence
of such temporal data, 
\citet{brandt2018querying} proposed \MTL{}---an extension
of Datalog~\cite{ceri1989you} with operators from
metric temporal logic~\cite{koymans1990specifying} interpreted
over the rational timeline. 
\MTL{} is a  powerful KR language,
which has found applications
in ontology-based query answering~\cite{brandt2018querying, kikot2018data,
guzel2018ontop, koopmann2019ontology} and stream
reasoning~\cite{walkega2019reasoning}. 
For example, the following \MTL{} rule can be used to analyse equipment data:
$$
\boxminus_{[0,1]} \mathit{ExcHeat}(x)  \gets  \boxminus_{[0,1]}\mathit{Temp24}(x) \land  \diamondminus_{[0,1]}\mathit{Temp41}(x).
$$
The rule states that a device $x$ has been under
excessive heat continuously within a past interval of length 1 ($\boxminus_{[0,1]}$) if  
the temperature recorded in this interval was always above 24 degrees ($\mathit{Temp24}$) and also, at some point in this interval ($\diamondminus_{[0,1]}$), the  temperature was above 41 degrees ($\mathit{Temp41}$).

Reasoning in \MTL{} is, however, of
high complexity, namely \EXPS-complete \cite{brandt2018querying} and
\PS-complete with respect to data size  \cite{walega2019datalogmtl},
which makes reasoning in practice challenging.
Thus, research has recently focused on 
establishing a suitable trade-off between expressive power and complexity of reasoning,
by 
identifying lower complexity fragments
of \MTL{} \cite{DBLP:conf/ijcai/WalegaGKK20,walega2021finitely} as well as studying 
alternative semantics with more favourable computational behaviour
\cite{DBLP:conf/kr/WalegaGKK20,ryzhikov2019data}.

The design and implementation of practical reasoning algorithms for the
full \MTL{} language, however, remains a largely unexplored area---something
that has  prevented its widespread adoption in applications.
In particular, the only implementation we are aware of  is the prototype 
by  \citet{brandt2018querying}, which is limited to
 non-recursive \MTL{} programs.
This is in stark contrast with  plain
Datalog,
for which a plethora of systems have been developed and successfully deployed
in practice
\cite{motik2014parallel,DBLP:conf/semweb/CarralDGJKU19,DBLP:journals/pvldb/BellomariniSG18}.

In this paper, we present the first practical reasoning algorithm for the full \MTL{} language,
which combines materialisation (a.k.a.\ forward chaining) and automata-based reasoning.
On the one hand, materialisation is the reasoning paradigm adopted in most  
Datalog systems \cite{DBLP:conf/rweb/BryEEFGLLPW07}; to check fact entailment in this setting, one first computes
 in a forward chaining manner
all facts logically entailed by
the input program  and dataset, 
and then verifies whether
the input fact
is included   
amongst the entailed facts. 
A direct implementation of materialisation-based
reasoning in \MTL{} is, however, problematic  since
forward chaining may require infinitely many
rounds of rule applications \cite{walega2021finitely}. 
On the other hand, 
\citet{walega2019datalogmtl}
introduced a decision procedure for \MTL{} which relies on 
 constructing
B\"{u}chi automata  and checking non-emptiness of their languages. This procedure has been
 introduced
for obtaining tight  complexity bounds,  and not with efficient
implementation in mind;
in particular, the constructed automata are of exponential size, 
which makes direct  implementations
impractical.

Our new algorithm deals with these difficulties by 
providing an effective way of combining the scalability of
materialisation-based reasoning and the completeness guaranteed by automata-based procedures, 
thus bringing together
`the best of both worlds'. To achieve this, our algorithm aims at minimising reasoning workload
by resorting to materialisation-based reasoning whenever possible while
minimising the use of automata.
 Furthermore,
we propose a suite of optimisation techniques aimed at
reducing the workload
involved in rule application during materialisation-based reasoning, as well as in B\"uchi automata
construction and the subsequent non-emptiness checks.

We have implemented our approach in a new reasoner called MeTeoR (\url{https://meteor.cs.ox.ac.uk}),
%\nb{P: I added  website address, but it is not working yet}%\footnote{The source code of the system and 
%a representative subset of the benchmarks are  
%provided as supplementary material. }
which supports fact entailment over arbitrary (i.e., potentially recursive) \MTL{} programs and large-scale
temporal datasets.
We have evaluated the performance of our reasoner on an extension of the
Lehigh University Benchmark~\cite{guo2005lubm} with temporal rules and data, 
as well as on a benchmark based on a real-world  meteorological dataset \cite{maurer2002long}.
Our results show that
MeTeoR is a scalable system that can
successfully reason over complex recursive programs and
datasets including tens of millions
of temporal facts;  furthermore, it consistently
outperformed the approach  by \citet{brandt2018querying} when
reasoning over non-recursive programs.

\section{Preliminaries}

We recapitulate the standard
definition of \MTL{}, interpreted with the standard continuous
semantics  over the rational timeline \cite{brandt2018querying}.

\paragraph{Syntax.}
A \emph{relational atom} is a function-free first-order atom of the form $P(\sbf)$,
with $P$ a predicate and $\sbf$ a tuple of \emph{terms}.
A \emph{metric atom} is an expression
given by the following grammar, where $P(\sbf)$ is a relational atom, and
$\diamondminus$, $\diamondplus$, $\boxminus$, $\boxplus$, $\Si$, $\Ui$ are MTL
operators indexed with positive rational intervals $\varrho$ (i.e.,  containing only non-negative rationals):
\begin{multline*}
	M  \Coloneqq
		\top  \mid  \bot  \mid
	P(\sbf)  \mid
	\diamondminus_\varrho M   \mid
	\diamondplus_\varrho M   \mid   \\
	\boxminus_\varrho M   \mid
	\boxplus_\varrho M  \mid
	M \Si_\varrho M   \mid
	M \Ui_\varrho M.
\end{multline*}
A \emph{rule} is an expression of the form of
\begin{equation}
\matA'  \leftarrow \matA_1 \land \dots \land \matA_n, \quad  \text{ for } n \geq 1, \label{eq:rule}
\end{equation}
with each $\matA_i$ a metric atom, and $\matA'$ a metric atom not
mentioning operators $\diamondminus$, $\diamondplus$, $\Si$, and $\Ui$; metric
atom $\matA'$ is the rule's \emph{head} and the conjunction $\matA_1 \land
\dots \land \matA_n$ is its \emph{body}. A rule is \emph{safe} if each variable
in its head also occurs in its body; a  \emph{program}
is a finite set of safe rules. An expression (metric atom, rule, etc.) is \emph{ground} if it mentions no variables.
 A \emph{fact} is an expression $\matA @ \varrho$ with $\matA$ a
ground relational atom and $\varrho$ a rational interval; a
\emph{dataset}  is a finite set of facts.
The \emph{coalescing} of facts $M@\varrho_1$ and $M@\varrho_2$, with
$\varrho_1$ and $\varrho_2$ intervals that are either adjacent or have
a non-empty intersection,
is the fact $M@\varrho_3$ where $\varrho_3$ is the union of $\varrho_1$
and $\varrho_2$. 
The \emph{grounding} 
$\ground{\Prog}{\D}$ 
of a program $\Prog$ with respect to a dataset $\D$ is a set of all ground rules  obtained 
by assigning constants from $\Prog$ and $\D$  to variables in  $\Prog$.

The \emph{dependency graph} of a  program $\Prog$ is the
directed graph $G_{\Prog}$,
with a vertex $v_P$ for each predicate $P$ in $\Prog$
and an edge $(v_Q, v_R)$ whenever there is a rule in $\Prog$ mentioning
$Q$ in the body and $R$ in the head.
Program $\Prog$ is \emph{recursive} if
$G_{\Prog}$ has a cycle.
A predicate $P$ is \emph{recursive} in   $\Prog$
if $G_{\Prog}$ has 
a path ending in $v_P$ and including
a cycle (the path can be a self-loop).
Furthermore, for a predicate $P$, a rule  $r$ is \emph{$P$-relevant} 
in $\Prog$ if 
there exists a rule $r'$ in $\Prog$ mentioning $P$ or $\bot$ in the
head and a path in $G_{\Prog}$ starting from 
a vertex representing the predicate in the head of $r$ and ending in a vertex representing some predicate 
from the  body  of $r'$.
Intuitively, $P$-relevant rules may be used for 
deriving facts about $P$ or $\bot$ (i.e., inconsistency).

\paragraph{Semantics.}
An \emph{interpretation} $\I$ specifies, for each ground relational atom $M$ and each time point $t \in \mathbb Q$, whether $M$
holds at $t$, in which case we write ${\mathfrak{I},t \models M}$.
This extends to metric atoms with MTL operators as shown in Table~\ref{semantics}.
\begin{table}[t]
\begin{alignat*}{3}
	&\I, t \models  \top    && && \text{for each } t
	\\
	&\I, t  \models \bot   && && \text{for no } t
	\\
	&\I,  t  \models \diamondminus_\varrho M    && \text{iff}   && \I, t' \models  M \text{ for some } t' \text{ with } t -   t' \in \varrho
	\\
	&\I,  t  \models  \diamondplus_\varrho  M  && \text{iff} &&  \I,  t' \models    M  \text{ for some } t'  \text{ with } t' - t \in \varrho
	\\
	&\I, t  \models \boxminus_\varrho  M  && \text{iff} && \I, t' \models M \text{ for all } t' \text{ with } t-t' \in \varrho
	\\
	&\I,t  \models \boxplus_\varrho  M   && \text{iff} && \I,t' \models  M \text{ for all } t'  \text{ with } t'-t \in \varrho
	\\
	&\I,t  \models M_1 \Si_\varrho M_2   &&  \text{iff} &&  \I,t' \models  M_2 \text{ for some } t'  \text{ with } t-t' \in \varrho
	\\
	& && && \text{and }  \I,t'' \models  M_1 \text{ for all } t'' \in (t',t)
	\\
	&\I,t  \models  M_1  \Ui_\varrho M_2 \;   &&  \text{iff} \; && \I, t' \models  M_2 \text{ for some } t'  \text{ with } t' - t \in \varrho
	\\
	& && && \text{and }  \I, t'' \models  M_1 \text{ for all } t'' \in (t,t')
\end{alignat*}
\caption{Semantics of ground metric atoms}
\label{semantics}
\end{table}
An interpretation $\I$ satisfies a  fact  $\matA @ \varrho$
if $\I,t \models \matA$ for all $t \in \varrho$. An interpretation $\I$ satisfies a ground rule $r$ if, whenever $\I$ satisfies each body atom of $r$ at a time point $t$, then $\I$ also satisfies the head of $r$ at $t$.
An interpretation $\I$ satisfies a (non-ground) rule $r$ if it satisfies each ground  instance of~$r$. An interpretation $\I$ is a \emph{model} of a program $\Prog$ if
it satisfies each rule in $\Prog$, and it is a \emph{model} of a dataset $\D$ if
it satisfies each fact in $\D$.
A program $\Prog$ and a dataset $\D$
are \emph{consistent} if they have a model, and they
\emph{entail} a fact $M@ \varrho$
if each  model of both $\Prog$ and $\D$ is also a model of $M@ \varrho$.
Each dataset $\D$ has the least model $\I_\D$, and we say that dataset $\D$ \emph{represents}  interpretation $\I_{\D}$. 

\paragraph{Canonical Interpretation.}
The \emph{immediate consequence operator} $T_{\Prog}$ for a
program $\Prog$ is a function mapping an
interpretation $\I$  to the least interpretation  containing
$\I$ and satisfying the following property for each
ground instance $r$ of a rule in $\Prog$: whenever 
$\I$ satisfies each body atom of $r$ at time point $t$, then
$T_{\Prog}(\I)$ satisfies the head of $r$ at $t$.
The successive application of $T_{\Prog}$ to $\I_\D$ defines
a transfinite sequence of interpretations $T_{\Prog}^{\alpha}(\I_\D)$ for ordinals~$\alpha$
as follows: (i)~${T_{\Prog}^0(\I_\D) = \I_\D}$, (ii)
${T_{\Prog}^{\alpha+1}(\I_\D) = T_{\Prog}(T_{\Prog}^{\alpha}(\I_\D))}$ for $\alpha$ an ordinal, and
(iii)~$T_{\Prog}^{\alpha} (\I_\D) = \bigcup_{\beta < \alpha} T_{\Prog}^{\beta}(\I_\D)$ for
$\alpha$ a limit ordinal.
The \emph{canonical interpretation} $\can{\Prog}{\D}$ of $\Prog$  and $\D$ is the interpretation
$T_{\Prog}^{\omega_1}(\I_\D)$, with
$\omega_1$ the first uncountable
ordinal.
If $\Prog$ and $\D$ have a model,
the canonical interpretation $\can{\Prog}{\D}$ 
is the least model of $\Prog$ and $\D$  \cite{brandt2018querying}.

\paragraph{Reasoning.}
The main reasoning tasks in DatalogMTL are \emph{fact entailment} and 
\emph{consistency checking}.
The former is to check  whether a program and dataset entail 
a given fact,  and the latter is to check whether a program and dataset are consistent.
These  problems are reducible to the complement of each other; both of them are \EXPS{}-complete \cite{brandt2018querying}
and  \PS{}-complete in data complexity (i.e. with respect to the size of a dataset) \cite{walega2019datalogmtl}.

%%%%%%%%
\section{Materialisation vs. Automata-based Reasoning}
%%%%%%%%

\emph{Materialisation}
(a.k.a. forward chaining) is a standard reasoning technique used in scalable  implementations of plain Datalog, which consists of
successive rounds of rule applications to compute all consequences of an input program and dataset
\cite{DBLP:journals/ai/MotikNPH19,DBLP:conf/rweb/BryEEFGLLPW07}.
The resulting set of facts
represents  the canonical model
over which all queries can be  answered directly.
As in plain Datalog, each consistent pair of a
\MTL{} program $\Prog$  and a dataset $\D$
admits
a canonical model $\can{\Prog}{\D}$ defined as
the least fixpoint of the
immediate consequence operator $T_{\Prog}$ capturing a single
round of rule applications. The use of metric operators in rules, however,
introduces two main practical difficulties.
First, \MTL{} interpretations are intrinsically infinite  and hence
a materialisation-based algorithm must be able to finitely represent
(e.g., as a dataset) the result of each individual round of rule applications.
Second,
reaching the least fixpoint of the
immediate consequence operator in \MTL{} may require an infinite number of rule
applications;
hence, a direct implementation of
materialisation-based reasoning is
non-terminating.
In contrast, materialisation is guaranteed to terminate for non-recursive programs, thus
 providing a decision procedure \cite{brandt2018querying}.

\citet{walega2019datalogmtl} provided an  automata-based decision procedure
for consistency checking applicable to
 unrestricted \MTL{} programs. This approach relies on
reducing consistency checking of an input
program $\Prog$ and dataset $\D$ to
checking non-emptiness of two
 B\"{u}chi automata.
These automata are responsible for accepting parts of a model located, respectively, to the left of the least number mentioned in $\D$, and to the right of the greatest number in $\D$.
Thus,  $\Prog$ and $\D$ have a model if and only if both
automata have non-empty languages.
States of these automata are polynomially representable in the size of $\D$,  so
the reduction yields a
\PS{} data complexity bound for consistency checking.
The idea behind the reduction is based on splitting a model
witnessing consistency of $\Prog$ and $\D$ into fragments
 corresponding to segments of the timeline;
automata states describe metric atoms holding
in such segments and the
transition functions  guarantee that atoms from successive fragments
do not violate the semantics of metric operators.
Although this automata-based approach provides tight complexity
bounds, it does not yield a practical decision procedure
since the automata contain a large (exponential in the size of $\D$) number of
states.

%In what follows we will introduce a new practical decision procedure which brings together the advantages of  materialisation and the automata-based approach.

\section{A Practical Decision Procedure}\label{sec:reasoning}

In this section, we present our approach to practical reasoning
in the full DatalogMTL language.
Our algorithm decides fact entailment, that is, checks whether a given fact  is entailed by a
given program and dataset.
To this end, we  optimise and combine
materialisation---which is a scalable technique, but is not guaranteed to terminate---,
and  the automata-based approach---which is terminating in general,
but less efficient in practice.
The key novelties of our approach are as follows:

\begin{itemize}
\item[--] an optimised implementation of the
materialisation approach, which aims at applying  the immediate consequence operator efficiently  and
storing a succinct representation of the fragment of the canonical interpretation constructed thus far;

\item[--] an optimised implementation of the automata-based reasoning approach of \citet{walega2019datalogmtl}, which aims at minimising the size of the automata; and

\item[--]  an effective way of combining materialisation with automata-based reasoning
 that aims at reducing reasoning workload and achieving early termination.
\end{itemize}
In the remainder of this section we will discuss each of these components in detail.

%%%%%%%%%%%%%
\subsection{Optimising Materialisation}

The key component of materialisation-based reasoning is an
effective implementation of the immediate consequence operator
capturing a single round of rule applications.
Our implementation takes as input a program $\Prog$ and a dataset $\D$
and computes a dataset $\D'$ representing the interpretation $T_\Prog (\I_\D)$ obtained by
application of $T_{\Prog}$ to the least
model of $\D$.
Thus, it addresses one of the difficulties
associated to materialisation-based reasoning by
ensuring that the result of a single round of rule applications is a  finite object.

To facilitate the presentation of our rule application procedure, we first
briefly discuss
 details regarding the representation and storage of datasets.
We associate to each ground relational atom
a list of intervals sorted by their left endpoints, which
provides a compact
account of all facts mentioning this ground relational atom.
Moreover, each ground relational atom is indexed by a composite key consisting of its
predicate and its tuple of constants. This layout is useful
for fact coalescing
and fact entailment checking;  for instance,
to check if fact  $M @ \varrho$ is entailed
by dataset $\D$,
it suffices to find $\varrho' $ such that
$M @ \varrho' \in \D$ and  $\varrho \subseteq \varrho'$; this can be achieved by
first scanning the sorted list of intervals for $M$
using the index and checking
if $\varrho$ is a subset of one of these intervals.
Additionally, each ground relational atom is also indexed by
each of its term arguments  to facilitate joins.
Finally, when a fact is inserted into the dataset,
the corresponding list of intervals  is sorted to
facilitate subsequent operations.

We perform a single round of rule applications with a procedure $\mathsf{ApplyRules}$ presented in  Algorithm~\ref{materialisation}.
To construct a representation of $T_\Prog(\I_D)$, for an input program $\Prog$ and a dataset $\D$, our algorithm performs two main steps.
The first step is performed for each
ground rule in  $\ground{\Prog}{\D}$ (Line~\ref{alg:onestep:ruleinstanceloopstarts}), where
 rule grounding is performed using
 a standard
index nested loop  join algorithm \cite{DBLP:books/daglib/0020812} since
facts in $\D$ are indexed.
Then, the first step consists of extending $\D$ with all facts that can be derived by applying $r$ to $\D$ (we also add $\bot$ to $\D$ if some rule leads to inconsistency).
Hence, the new dataset $\D'$ represents the interpretation $T_{\{ r\}} (\I_D)$ (Line~\ref{alg:onestep:evaluate}).
To implement the first step, we modified the completion rules for normalised  programs defined by
\citet[Section 3]{brandt2018querying}---our approach is suitable for arbitrary \MTL{} rules
and thus not only for rules in the aforementioned normal form, which disallows nested MTL operators,
the use of
$\diamondplus$ and $\diamondminus$ in rule bodies, and the use of $\boxplus$ and $\boxminus$ in rule heads \cite{brandt2018querying}.
Since our approach does not
require program normalisation
as a pre-processing step,  we avoid the computation and storage
of auxiliary predicates introduced by the
normalisation.
Furthermore we implement an optimised version of (temporal) joins that is required to evaluate rules with several body atoms.
A na\"ive implementation of the join of
metric atoms $M_1, \dots, M_n$ occurring in the body of a rule would require computing all intersections of intervals $\varrho_1, \dots, \varrho_n$ such that $M_i @ \varrho_i$ occurs in the so-far constructed dataset, for each $i \in \{1, \dots,  n\}$.
Since each $M_i$ may hold in multiple intervals in the dataset, the na\"ive approach is ineffective in practice.
In contrast, we implemented
a variant of the sort-merge join algorithm: we first sort all $n$  lists of intervals corresponding to $M_1, \dots, M_n$, and
then  scan the sorted lists to compute the intersections, which improves performance.
Our approach to temporal joins can  be seen as
 a  generalisation of the idea
sketched by \citet[Section 5]{brandt2018querying}, which deals with two metric atoms at a time but has not
been put into practice.

% \nb{P: I added Line 1, in Line 3 changed $\D$ to $\D'$, then in Lines 4-8 I changed all $\D$ to $\D'$. }
% \nbM{Hu: Line 3 seems confusing: shouldn't we always add more facts to D'?}
% \nb{P: I changed Line 3, is it OK now?}

%\nb{P: Line 2 in the algorithm seems wrong, as now the algorithm modifies in in each step $\D$ occurring in $T_{\{ r \}}(\I_\D)$. I believe that this $\D$ should be always the same, namely it should be the input dataset.}

%\nb{\textcolor{black}{Dingmin: Line 2 should be changed to $\D' \coloneqq$ representation of $T_{\{ r \}}(\I_\D) \cup \D'$; Accordingly, the $\D$ in line 3-7 should be changed to $\D'$}}

\begin{algorithm}[th]
    \SetKwInput{Input}{Input}
    \SetKwInput{Output}{Output}
    \Input{a program $\Prog$ and a dataset $\D$}
	 \Output{a dataset}
        $\D' \coloneqq \D;$
        	 
        \For{each rule $r \in \ground{\Prog}{\D}$}{ \label{alg:onestep:ruleinstanceloopstarts}
                  $\D' \coloneqq$ representation of $\I_{\D'} \cup T_{\{ r \}}(\I_\D)$;
                                                                    \label{alg:onestep:evaluate}

            % \While{some facts in $\D$ can be coalesced}{
            %
            % Take $M\! @ \! \varrho, M \! @ \! \varrho' \in\D$ that can be coalesced;
            %
            % Replace  $M \! @ \! \varrho$ and $M \! @ \! \varrho'$ in $\D$ with $M \! @ \! \varrho \! \cup \! \varrho' \!$;
            %
            % }
        }                                                                                             \label{alg:onestep:ruleinstanceloopends}
		\For{each ground metric atom $M$ appearing in $\D'$}{
		    Retrieve all $M   @   \varrho_1, \dots, M   @  \varrho_n$ occuring in $\D'$;  \\
			Merge $\varrho_1, \dots, \varrho_n$ into $\varrho'_1, \dots, \varrho'_m$; \\
			Replace $M  @   \varrho_1, \dots, M  @  \varrho_n$ with $M  @  \varrho'_1, \dots, M  @  \varrho'_m$ in $\D'$;
		}
        Return $\D'$;                                                                    \label{alg:onestep:coalesce}
    \caption{$\mathsf{ApplyRules}$}\label{materialisation}
\end{algorithm}

% The second step of Algorithm \ref{materialisation}
% consists of coalescing facts in the so-far constructed dataset until no more facts can be coalesced (Lines 3-5).

The second step of Algorithm \ref{materialisation}
consists of coalescing facts in the so-far constructed dataset (Lines 4--7).
Conceptually, whenever there exist facts $M @ \varrho $ and $M @ \varrho' $  in $\D'$
%\nb{P: new} 
%the constructed representation of $T_{\{ r \}}(\I_\D)$ 
such that
$\varrho$ and $\varrho'$ can be coalesced, we replace them
with ${M @ \varrho \cup \varrho'}$.
% and repeat this
%process until no more facts can be coalesced.
To achieve this in practice, for each ground relational
atom occurring in $\D'$
%\nb{P: new} 
 we iterate
through the corresponding
sorted list of intervals and merge them
as needed (Lines 5--6). Replacing intervals with coalesced ones  reduces memory usage while
preventing redundant computations in subsequent rounds of
rule applications.
Finally, in Line~8,  Algorithm \ref{materialisation} returns a  dataset, which represents  $T_\Prog(\I_D)$.

%%%%%%%%

%%%%%%%%
\subsection{Optimising Automata-based Reasoning}

We have implemented an optimised variant of
the automata-based approach for 
checking consistency of a program $\Prog$ and dataset $\D$
introduced by \citet{walega2019datalogmtl}, which
is based on verifying non-emptiness of two  B\"{u}chi automata.
States of these automata are of size
polynomial in  $\D$
and exponential in the combined size of $\Prog$ and $\D$, 
which makes automata construction and the 
non-emptiness checks hard in practice.
To address this difficulty, our implementation
introduces a number of 
optimisations and we next highlight two of them.
% Our implementation takes as input a program $\Prog$, a dataset $\D_1$,
% and a 'partially materialised' dataset $\D_2$ containing facts entailed by $\Prog$
% and $\D_1$; the purpose of the latter dataset will become clear later on.

First, instead of directly checking 
consistency of $\Prog$ and $\D$, our implementation
checks consistency of $\Prog$ and the `relevant' part of $\D$, namely the subset
$\D'$ of facts in $\D$ 
mentioning predicates occurring in the bodies of rules in $\Prog$.
This optimisation is based on the straightforward
observation that  $\Prog$ and $\D$ are consistent if and only if $\Prog$ and $\D'$ are.
In practice, $\D'$ can be significantly 
smaller than $\D$, with the subsequent reduction in the size of
the constructed automata.
%Since our implementation  uses $\D'$ instead of $\D$, the performance of the approach increases.

%Second, we divide $\D'$ into a set $\D'_\mathit{nrec}$ of facts mentioning predicates that are non-recursive in $\Prog$ 
%and a set $\D'_\mathit{rec}$ of facts mentioning recursive predicates  in $\Prog$.
%We then materialise $\Prog$ and $\D'_\mathit{nrec}$, which is guaranteed to 
%terminate since all predicates involved 
%in the data are non-recursive in $\Prog$.
%This materialisation allows us to determine exactly at which time
%points these 
%non-recursive predicates
%hold in  the canonical model of $\Prog$ and $\D'$.
%This information is very valuable for optimising
%the  non-emptiness for the constructed automata; in particular, when searching for accepting runs of the
 %automata we can reduce the search space by considering only
%states that match the constructed materialisation of $\Prog$ and $\D'_\mathit{nrec}$ on facts mentioning 
%non-recursive predicates in $\Prog$.

Second, 
we have optimised the construction of  states of the 
automata when searching for accepting runs.
Since automata states represent fragments of a model, we can exploit 
the semantics of MTL operators to restrict possible locations of metric atoms holding in the same state.
For example, if $\boxplus_{[0,\infty)}P$ holds at a time point $t$, then it needs to hold in all time points greater than $t$;
similarly, if $\diamondplus_{[0,\infty)}P$ holds at $t$, then it needs to hold at all time points smaller than $t$, and analogous statements  hold for metric atoms with past operators $\boxminus$ and $\diamondminus$.
We have incorporated a suite of such  restrictions, which resulted in a more efficient construction of automata states.
\subsection{Combining Materialisation and Automata}
%%%%%%

Our approach
aims at minimising the reasoning workload by
considering only
relevant parts of the input
during reasoning, and by resorting to 
materialisation-based reasoning
whenever possible while minimising the use of automata.
  
Given as input a program $\Prog$,  a dataset $\D$, and 
a fact $P(\sbf)@\varrho$, our procedure returns a truth value stating whether $\Prog$ and $\D$ entail $P(\sbf)@\varrho$.
To this end, we proceed as  summarised in
Algorithm~\ref{alg:pipline}.
The algorithm starts by checking whether $P(\sbf)@\varrho$ is already
entailed by $\D$ alone, in which case  $\mathsf{True}$ is returned in Line 1.
%\nb{P: Pan, can you please check the numbering of Lines? Reviewer found some mistakes, and U believe that you agreed with him.}
As already discussed, this 
check can be 
realised very efficiently using suitable indexes.

\begin{algorithm}[th]
  % \DontPrintSemicolon
\SetKwInput{Input}{Input}
\SetKwInput{Output}{Output}
\SetKwFor{Loop}{loop}{}{}
\SetKwFor{threadone}{thread 1:}{}{}
\SetKwFor{threadtwo}{thread 2:}{}{}
  \Input{A program $\Prog$, a dataset $\D$, and a fact $P(\sbf)@\varrho$}
  \Output{A truth value}

  \BlankLine

\textbf{if} $\D \models P(\sbf)@\varrho$ \textbf{then}  Return $\mathsf{True}$; \\

 $\Prog^{P} \coloneqq$ the set of  all $P$-relevant  rules in $\Prog$;

  \If{$\Prog^{P}$ is non-recursive}{
  \Loop{}{                                        \label{alg:pipeline:nonrec:whilestarts}
       % $\D' \leftarrow \D$ \\
        $\D' \coloneqq \mathsf{ApplyRules}(\Prog^P, \D)$; \\
        \lIf{$\D'  \models P(\sbf) @ \varrho$}{Return $\mathsf{True}$} 
        \lIf{$\D' = \D$}{Return $\mathsf{False}$}
        $\D \coloneqq \D'$;
        }
     }                                                           \label{alg:pipeline:nonrec:whileends}
 \Else{

     %  $\D^{orig} \leftarrow \D$ \\
       \Loop{}{
           $\D' \coloneqq \mathsf{ApplyRules}(\Prog^P, \D);$ \\

           \lIf{$\D' \models  P(\sbf)@\varrho$}{Return $\mathsf{True}$} 

           \lIf{$\D' = \D$}{Return $\mathsf{False}$}
            $\D \coloneqq \D'$; \\
           \textbf{if} all facts with non-recursive predicates are derived \textbf{then} \textbf{break};

       }

       $\D^{\mathit{pre}} \coloneqq \D$; \\
       run two threads in parallel:\\

       \textbf{Thread 1:} \\
       %\threadone{}{
        \Loop{}{                                               \label{alg:pipeline:rec:whilestarts}
       % $\D' \leftarrow \D$ \\
        $\D' \coloneqq \mathsf{ApplyRules}(\Prog^P, \D)$; \\
        \lIf{$\D' \models  P(\sbf)@\varrho$}{ Return $\mathsf{True}$}
        \lIf{$\D' = \D$}{Return $\mathsf{False}$}
        $\D \coloneqq \D'$;
        }
        %}
     \textbf{Thread 2:} \\
	%\threadtwo{}{
     $\Prog', \D{'} \coloneqq \mathsf{EntailToInconsist}(\Prog^P, \D^{\mathit{pre}}, P(\sbf)@\varrho)$;  \\
     Return negation of  $\mathsf{AutomataProcedure}(\Prog',\D{'})$;
%     \If{ $\mathsf{AutomataProcedure}(\Prog',\D{'})$ returns $\mathsf{True}$}{Return $\mathsf{False}$;}
%     Return $\mathsf{True}$;
	}
     %}

  \caption{Practical fact entailment}
  \label{alg:pipline}
\end{algorithm}

In Line 2, we identify (using the program's dependency graph $G_\Prog$) the subset $\Prog^P$ of $P$-relevant rules in $\Prog$, that can be potentially used to derive $P(\sbf)@\varrho$ (or $\bot$, i.e.,  inconsistency). This  allows us to disregard a potentially
large number of irrelevant rules during reasoning, optimise automata construction, and also
identify cases where materialisation
naturally terminates and automata-based reasoning
is not required.
In particular, if $\Prog^P$ is non-recursive, fact entailment can be
decided using only materialisation (Lines 3--8), namely rules of $\Prog^P$ are applied (Line 5) until entailment of $P(\sbf)@\varrho$  is detected (Line 6) or a fixpoint of the immediate consequence operator is reached (Line 7).

Even if $\Prog^P$ is recursive,  
we can still benefit from  using materialisation.
For this, we start by
performing a `pre-materialisation' step (Lines 10--16), where
we materialise until
all facts mentioning non-recursive predicates in $\Prog^{P}$ have been derived, in which case we break the 
pre-materialisation loop and continue to the next stage.
Upon completion of the pre-materialisation step, we 
run two threads in parallel, where the
first thread continues applying materialisation on a best-effort basis (Lines 
18--23) and the second thread resorts to automata (Lines 24--26); 
the algorithm terminates as soon as one of the threads generates an output truth value.

The second thread
reduces fact entailment  to inconsistency of a new program $\Prog'$ and 
dataset $\D'$ using the  reduction from
the literature \cite{brandt2018querying,walega2019datalogmtl} and
then runs the automata-based
procedure for consistency checking described in the previous section.
Our reduction from entailment to inconsistency (Line 25)
 uses the set of relevant
rules  $\Prog^P$ instead of $\Prog$, 
 which reduces
the size of the automata; 
furthermore we use the pre-materialised dataset $\D^{\mathit{pre}}$ instead of the input dataset $\D$ to optimise the
 non-emptiness tests for the constructed automata.
Indeed,  with $\D^{\mathit{pre}}$ we can determine
entailment of all facts with non-recursive predicates, and
this allows us to reduce the search space
when checking existence of accepting runs of the automata.

Algorithm~\ref{alg:pipline} is clearly sound and complete. 
Furthermore, it is designed so that, on the one hand, the majority of entailment tests are handled via
 materialisation and, on the other hand,  the 
recursive program passed on as input to the
automata-based approach
is small. These features of the algorithm
will be empirically confirmed by our experiments described in the
following section.

%%%%

\section{Implementation and Evaluation}

\begin{table*}[th]
    \centering
\setlength{\tabcolsep}{0.3pt}
\renewcommand{\arraystretch}{1.2}
    \begin{tabular}{l|c|c|c|c|c|}
          & $\mathbf{T_1}$  &  $\mathbf{T_2}$  &  $\mathbf{T_3}$  & $\mathbf{T_4}$  & $\mathbf{T_5}$ \\
         \cline{2-5}
         \hline
         $\D_L^1$ &  $0.6$s($c=0.6$s,$n=0$)  &$70.1$s($c=2.1$s,$n=1.7$)    &$66.9$s($c=4.1$s,$n=1.9$)    &$97.8$s($c=3.2$s,$n=21.4$)    & $7.6$min($p=2.7$s)\\
         \hline
         $\D_L^2$ &  $0.8$s($c=0.8$s,$n=0$) &$152.5$s($c=8.9$s,$n=1.5$)   &$167.9$s($c=11.2$s,$n=3.1$) &$229.3$s($c=10.2$s,$n=17.1$) & $50.2$min($p=10.5$s)\\
          \hline
         $\D_L^3$ &  $1.9$s($c=1.9$s,$n=0$) &$201.4$s($c=13.1$s,$n=1.6$)  &$269.2$s($c=18.3$s,$n=2.9$) &$284.8$s($c=27.3$s,$n=11.1$) &$92.1$min ($p=19.7$s) \\
          \hline
         $\D_L^4$ &  $2.5$s($c=2.5$s,$n=0$) &$301.2$s($c=21.3$s,$n=1.8$)  &$341.9$s($c=26.1$s,$n=2.6$) &$312.5$s($c=41.3$s,$n=7.8$)   &$172.6$min ($p=28.6$s) \\
          \hline
    \end{tabular}
    \caption{Results of Experiment 1a}
    \label{tab:recursive}
\end{table*}

We have implemented a
 prototype \MTL{} reasoner, called MeTeoR, which
 exploits our practical decision procedure from Algorithm \ref{alg:pipline} to decide fact entailment.
Our implementation uses existing libraries in the Python 3.8 eco-system without depending on other third-party libraries.
To the best of our knowledge, no reasoning benchmark is currently available
for \MTL{}\footnote{The datasets used by \citet{brandt2018querying} to test their reasoning approach for non-recursive programs are not publicly available.}, so to evaluate the
performance of MeTeoR we have
designed two new benchmarks.

\paragraph{LUBM Benchmark.}
We obtain the first benchmark  by extending
the Lehigh University Benchmark (LUBM)~\cite{guo2005lubm} with temporal rules and data.
To construct temporal datasets we modified LUBM's data generator,
which provides means for
generating datasets of different sizes,
to randomly assign an interval to
each generated fact; the
rational endpoints of each interval belong to a  range, which is a customised parameter.
We used the same approach to  generate  input query facts for entailment checks.
In addition, we extended
the $56$ plain Datalog rules obtained from the
OWL 2 RL fragment of LUBM with
$29$
temporal rules involving recursion and mentioning all
metric operators available in \MTL{}.
We denote the resulting \MTL{} program with $85$ rules as
$\Prog_L$.

\paragraph{Meteorological Benchmark.}
For this benchmark,
we used a freely available
dataset with meteorological observations \cite{maurer2002long};
in particular, we used a set $\D_W$ of 397 millions of facts from the years 1949--2010.
We then adapted the  \MTL{} program used by \citet{brandt2018querying} to reason with \MTL{} about weather, which resulted in a
a non-recursive program $\Prog_W$
with 4 rules.

\paragraph{Machine Configuration.}
All experiments have been conducted
on a Dell PowerEdge R730 server with 512 GB RAM and two Intel Xeon E5-2640 2.6 GHz processors running Fedora 33, kernel version 5.8.17.
%We have performed two experiments for each benchmark.\nb{P: this seems not not a true statement?}
Our first experiment tests the scalability
of MeTeoR; the second experiment compares the
performance of MeTeoR and the approach of \citet{brandt2018querying}
when reasoning over non-recursive programs.
In both experiments, each test was conducted once.

\subsection{Experiment 1:  Scalability of MeTeoR}

%\smallskip
%\noindent\textbf{Scalability of MeTeoR}
In Experiment 1a, to evaluate the scalability of MeTeoR, we used the recursive
program $\Prog_L$  from our
LUBM benchmark and four datasets (obtained with LUBM's data generator) $\D_L^1$, $\D_L^2$, $\D_L^3$, and $\D_L^4$ of increasing size, which
consist of  $5$, $10$, $15$, and  $20$ million facts, respectively.
As already mentioned, input query facts
are randomly generated.
We hypothesise that
performance
critically depends on the type of these input query facts and, in particular,
on which block within Algorithm \ref{alg:pipline} the answer is returned.

To check our hypothesis
we have identified five disjoint types of input query facts
for an input program $\Prog$ and dataset $\D$,
depending on where in the execution of
 Algorithm \ref{alg:pipline}  
 an answer is returned.
%$\mathbf{T_1}$, $\mathbf{T_2}$, $\mathbf{T_3}$, $\mathbf{T_4}$, and $\mathbf{T_5}$.
In particular, Algorithm \ref{alg:pipline} terminates in
\begin{itemize}
\item[--]
Line 1 for facts of type $\mathbf{T_1}$, so these facts 
are already entailed by the input dataset $\D$;
\item[--] Lines 6 or 7 for
facts of type $\mathbf{T_2}$, so  these facts 
are not entailed by $\D$ and mention a predicate whose relevant
rules in $\Prog$ are non-recursive;
\item[--] Line 13 or 22 for facts of type 
$\mathbf{T_3}$, so  materialisation of the relevant (recursive) subprogram 
reaches a fixpoint,
and the fact is not entailed by the resulting materialisation;
\item[--] Line 12 or 21 for facts  of type 
$\mathbf{T_4}$, so these facts have a recursive relevant subprogram and
are found to be entailed after finitely many rounds of rule application; and
\item[--]
 Line 26 for facts of type $\mathbf{T_5}$, so the entailment of these
  facts is checked using 
automata.
 \end{itemize}

The results of Experiment 1a are reported in Table~\ref{tab:recursive},
where we have generated 10 facts of each type and recorded the average runtime
of MeTeoR for such sets of 10 facts;
additionally, we have stated in brackets how much time was  consumed for fact coalescing ($c$) and how many rounds of rule applications ($n$) were performed  on average.
In the case of type $\mathbf{T_5}$ we report, instead, the time consumed by the pre-materialisation step ($p$).
The experiment shows that the performance of MeTeoR is dependent
on the type of an input fact, and in particular, on whether the system can verify it using
materialisation only (types $\mathbf{T_1}$--$\mathbf{T_4}$)  or automata are needed (type $\mathbf{T_5}$).
As expected, runtimes increase with the size of the
dataset and the number of rounds of rule applications needed.
Furthermore, although materialisation was more efficient, it is worth noting that
the automata-based approach could also be successfully applied
to such
large-scale datasets.
Finally, coalescing times were relatively small, and so were
pre-materialisation times for  facts of type $\mathbf{T_5}$.

If the majority of input facts can be
verified using materialisation only, then we can expect
robust and scalable performance; we hypothesise that this is likely to be the case
in many practical situations.
To verify this hypothesis on our benchmark, we have  generated $1,000$ random facts for
each of the datasets $\D_L^1$--$\D_L^4$  and
calculated the percentage of
facts that  belong to each of the types
$\mathbf{T_1}$--$\mathbf{T_5}$ (with respect to the considered dataset and the program $\Prog_L$).
We found that in $46.0\%$ cases the facts
were of type $\mathbf{T_1}$, $28.3\%$ of type
$\mathbf{T_2}$, $16.6\%$ of type $\mathbf{T_3}$, $8.3\%$ of type $\mathbf{T_4}$,
and  only $0.8\%$ of type $\mathbf{T_5}$.
This supports
our hypothesis that input facts requiring the use of automata
may rarely occur in practice.

\begin{figure}[ht]
\centering
\begin{tikzpicture}
\pgfplotsset{%
    width=0.45\textwidth,
    height=0.26\textwidth
}
        \begin{axis}[
            xlabel=Size of a dataset (in millions),
            ylabel=Run time (in seconds),
            xmin=50, xmax=160,
            ymin=0, ymax=3000,
            xtick={50,70, 90, 110, 130,150},
            xticklabels={$50$, $70$, $90$, $110$, $130$, $150$},
						ytick={0, 1000, 2000, 3000},
						yticklabels={$0$,  $1.000$,  $2.000$,  $3.000$},
            legend style={at={(axis cs:54,1390)},anchor=south west}
                    ]
            \addplot[color=red,mark=x, thick] plot coordinates {
									(50, 641.12)
									(70, 798.89)
									(90, 1048.29)
									(110,1419.07)
									(130,1798.78)
									(150,2489.82)

            };
            \addlegendentry{T2}

            \addplot[mark=*,blue, thick]
                plot coordinates {
									(50,441.66)
									(70,898.33)
									(90,1398.82)
									(110,1891.21)
									(130,2281.34)
									(150,2781.04)
                };
            \addlegendentry{T3}

            \addplot[mark=triangle, green!60!black, thick]
                plot coordinates {
									(50, 202.16)
									(70, 412.73)
									(90, 821.32)
									(110,1491.41)
									(130,1521.84)
									(150,2271.04)
                };
            \addlegendentry{T4}

         \end{axis}

    \end{tikzpicture}
 \caption{Results of Experiment 1b}
 \label{fig:crash}
\end{figure}
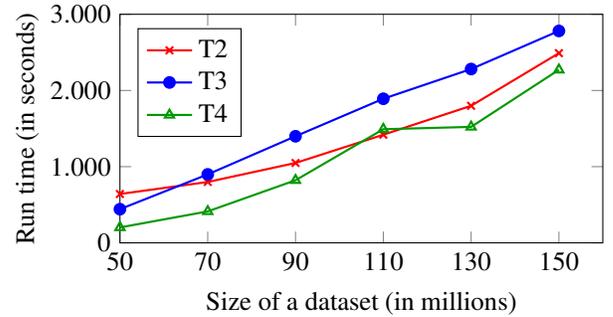

Finally, in Experiment 1b, we
 have stress tested MeTeoR.
To this end, we proceeded as in Experiment 1a, but
we keep constructing bigger datasets, until the average run times exceed 40 minutes.
Note that this threshold is already exceeded by facts of type $\mathbf{T5}$ in Experiment 1a, whereas facts of type $\mathbf{T1}$ do not require rule applications.
Therefore, in Experiment 1b we focused on types $\mathbf{T2}$, $\mathbf{T3}$, and $\mathbf{T4}$.
Our results, which are shown in
Figure~\ref{fig:crash},
 suggest that MeTeoR scales very well and it is capable of handling datasets
containing up to 150 million temporal facts using materialisation.

%  for facts of types
%$\mathbf{T_2}$ and $\mathbf{T_3}$---the most common
%types of facts requiring materialisation-based reasoning only
%We again generated
%datasets of increasing sizes,
%used MeTeoR to check fact entailment,
% and reported the average running times.
% When three or more facts out of ten of a given type had runtimes higher than 30 minutes, we deemed that the `limit' was reached and stopped generating larger datasets.
%Our results, which are shown in
%Figure~\ref{fig:crash},
% suggest that MeTeoR scales very well and it is capable of handling datasets
%containing up to 150 million temporal facts using materialisation.

\subsection{Experiment 2: Comparison with Baseline}
%Approach on Non-recursive Programs}

We performed two experiments to compare the performance of
MeTeoR and the baseline approach of \citet{brandt2018querying} for reasoning with non-recursive programs.
The approach by \citet{brandt2018querying} is based
on query rewriting----given a target predicate $P$ and an input program $\Prog$ the
algorithm generates a SQL query that, when evaluated over the input dataset $\D$,
provides the set of all facts with maximal intervals
over $P$  entailed by $\Prog$ and $\D$.
%  the
% input program and dataset.  and
% evaluates them over the input  dataset.
% The approach generates all facts
% which are entailed by the input program and dataset, and
% which mention the target predicate and maximal possible intervals.
To the best of our knowledge there is no
publicly available implementation of this approach; thus, we have produced our
own implementation. Following~\citet{brandt2018querying}, we used temporary tables (instead of subqueries)
to compute the extensions of predicates appearing in $\Prog$ on the fly, which has
two implications. First, we usually have not just one SQL query but a set of
queries for computing the final result; second, similarly to MeTeoR,
the approach
essentially works bottom-up rather than top-down.
The implementation by \citet{brandt2018querying} provides two
variants for coalescing: the first one
uses standard SQL queries by
\citet{zhou2006efficient}, whereas the second one implements coalescing
explicitly.
For our baseline we adopted the standard SQL approach, which
is less dependent on implementation details. Finally, we used Postgresql 10.18 for
all our baseline experiments.
In each test, we could verify that
the answers returned by MeTeoR and the baseline  coincided.

In Experiment 2a, we
compared the performance of our system with that of the baseline
using three
target predicates  from our LUBM
benchmark program $\Prog_L$, whose relevant sets of
rules constitute non-recursive programs $\Prog_L^1$, $\Prog_L^2$, and
$\Prog_L^3$ consisting of $5$, $10$, and $21$ rules, respectively.
For each program $\Prog_L^1$--$\Prog_L^3$ and each dataset $\D_L^1$--$\D_L^4$ from our benchmark, we used the baseline approach and MeTeoR to
compute all facts over the target predicates
entailed by the corresponding program and dataset---this was
achieved by MeTeoR using materialisation only since these programs
 are non-recursive. As a result,
 MeTeoR computed a representation of the entire
 canonical model, and not just the extension of the target predicates.
 As shown in Table~\ref{tab:materialise}, both approaches exhibit
good performance and scalable behaviour as the size of the data increases, with
our system consistently outperforming the baseline.

% \begin{table}[th]
%     \centering
%     \begin{tabular}{l|c|c|c|c|c|}
%          \multicolumn{2}{c|}{} & $\D_1$ &  $\D_2$ &  $\D_3$  &  $\D_4$ \\
%           \hline
%           \multirow{2}{*}{$\Prog_1$}  & baseline&  $160.3$s & $331.8$s & $503.2$s & $683.0$s  \\
%            & MeTeoR  & $51.3$s	& $90.5$s	& $114.15$s	& $170.3$  \\
%             \cline{2-6}
%             \hline
%           \multirow{2}{*}{$\Prog_2$} & baseline &
%              $345.9$s & $747.8$s & $1154.0$s & $1575.6$s \\
%             & MeTeoR  &  $146.8$ & $321.9$ & $425.8$ &$582.7$s \\
%             \cline{2-6}
%           \hline
%            \multirow{2}{*}{$\Prog_3$ } & baseline &  $523.9$s & $1118.5$s & $1688.5$s & $2254.6$s \\
%           & MeTeoR &  $213.4$s &	$455.2$s	& $760.1$s& $953.4$s\\
%             \cline{2-6}
%           \hline
%     \end{tabular}
%     \caption{Results of Experiment 2a
% %    Comparison of MeTeoR and our baseline on non-recursive programs. To ensure fairness,
% %    reported times do not include data loading times in our approach, nor the time needed to load data
% %    into the Postgres datatabase in the baseline.
%     }
%     \label{tab:materialise}
% \end{table}

\begin{table}[th]
    \centering
\setlength{\tabcolsep}{0.9pt}
\renewcommand{\arraystretch}{1.2}
    \begin{tabular}{l|c|c||c|c||c|c|}
         \multirow{2}{*}{}& \multicolumn{2}{c||}{$\Prog_L^1$} &
         \multicolumn{2}{c||}{$\Prog_L^2$} &
         \multicolumn{2}{c|}{$\Prog_L^3$}\\
         \cline{2-7}
         & baseline & MeTeoR & baseline & MeTeoR & baseline & MeTeoR  \\
         \hline
         $\D_L^{1}$ &  $160.3$s & $51.3$s  & $345.9$s  &  $146.8$s  & $523.9$s &  $213.4$s\\
         \hline
         $\D_L^{2}$ &  $331.8$s & $90.5$s &  $747.8$s   & $321.9$s & $1118.5$s&	$455.2$s\\
          \hline
         $\D_L^{3}$ &  $503.2$s & $114.2$s	 & $1154.0$s  & $425.8$s& $1688.5$s	& $760.1$s \\
          \hline
         $\D_L^{4}$ &  $683.0$s & $170.3$ & $1575.6$s &$582.7$s& $2254.6$s& $953.4$s \\
          \hline
    \end{tabular}
    \caption{Results of Experiment 2a}
    \label{tab:materialise}
\end{table}

In Experiment 2b, we performed similar tests as in Experiment 2a, but  using the meteorological benchmark.
We chose two target predicates from the program $\Prog_W$ which correspond to non-recursive
programs $\Prog_W^1$ and $\Prog_W^2$, each with two rules. In addition,
we constructed  subsets $\D_W^i$ of the entire meteorological dataset $\D_W$, where
the superscript $i$ indicates the number of years covered by the  dataset; in particular  $\D_W^{62}$ is the full dataset covering all 62 years, so $\D_W^{62}=\D_W$.
The results, as presented in Table~\ref{tab:weather}, show that
MeTeoR consistently outperforms the baseline approach.
%on $\Prog_W^1$. In the case of $\Prog_5$,
%although MeTeoR is slightly slower than the baseline for $\D_{10}$ and
%${\D_{20}}$, it tends to scale better when the size of the dataset increases.\nb{P: false statement}

\begin{table}[th]
   \centering
\renewcommand{\arraystretch}{1.2}   
   \begin{tabular}{l|c|c||c|c|}
        \multirow{2}{*}{}& \multicolumn{2}{c||}{$\Prog_W^1$} &
        \multicolumn{2}{c|}{$\Prog_W^2$} \\
        \cline{2-5}
        & baseline & MeTeoR & baseline & MeTeoR  \\
        \hline
        $\D_W^{10}$ &  $376.9$s & $141.8$s  & $38.8$s  &  $34.8$s \\
        \hline
        $\D_W^{20}$ &  $761.3$s & $324.0$s &  $78.5$s  & $77.9$s\\
         \hline
        $\D_W^{30}$ &  $1059.5$s & $430.8$s & $119.4$s  & $101.8$s \\
         \hline
        $\D_W^{40}$ &  $1406.5$s & $ 782.8$s & $163.23$s & $140.6$s \\
         \hline
        $\D_W^{50}$ &  $1765.1$s & $929.2$s & $203.7$s & $171.8$s \\
         \hline
        $\D_W^{62}$ &  $2234.0$s & $1050.8$s & $258.5$s & $240.4$s \\
         \hline
   \end{tabular}
   \caption{Results of Experiment 2b}
   \label{tab:weather}
\end{table}

Although in Experiments 2a and 2b the performance
of the baseline was comparable to
that of MeTeoR,
we
hypothesised that the performance gap would significantly increase (in favour of MeTeoR)
as the number of temporal joins involved during reasoning increases.
This is because computing temporal joins requires intersecting
 intervals---something for which databases do not seem to be
 optimised.
% and which
%would typically require quadratic time in the number of relevant intervals.\nb{P: quadratic if there are two conjuncts in a body, but in general it is a polynomial?}
In contrast, we anticipate that the performance of MeTeoR is more robust due to the way we implement
 joins (by first sorting  the sets of intervals involved in a join
%(in $O(nlogn)$)
and then linearly scanning these sets to compute the relevant intersections).

To verify this hypothesis we performed Experiment 2c which is similar to
Experiments 2a and 2b, but we considered only one target predicate  from the LUBM benchmark
with the corresponding (non-recursive) program $\Prog_L^1$, and  datasets $\D_L^{200}$,
$\D_L^{1000}$,  $\D_L^{2000}$, $\D_L^{3000}$, and $\D_L^{4000}$, which, for each relational atom, contain
increasing number of facts. In particular, these datasets contain facts
mentioning $7,082$ relational atoms---$\D_L^{200}$ has $200$ facts for
each of these atoms (so in total $1,416,400$ facts); whereas $\D_L^{1000}$, $\D_L^{2000}$, $\D_L^{3000}$, and $\D_L^{4000}$
have $1000$, $2000$, $3000$, and $4000$ facts, respectively. In general,
the larger is the number of facts mentioning the same relational atom, the
larger the number of temporal joins involved in reasoning. Thus, our
hypothesis is that, as we progress from $\D_L^{200}$ towards
$\D_L^{4000}$, we will see a larger performance gap between baseline and MeTeoR.

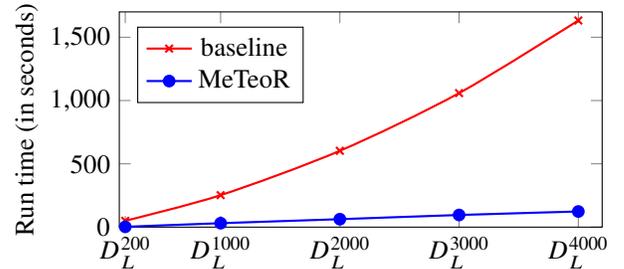
\begin{figure}[ht]
 \centering

% \resizebox{0.86\columnwidth}{!}{%
     \begin{tikzpicture}
\pgfplotsset{%
    width=0.45\textwidth,
    height=0.25\textwidth
}
        \begin{axis}[
            ylabel=Run time (in seconds),
            xmin=150, xmax=4200,
            ymin=0, ymax=1700,
            xtick={200, 1000, 2000, 3000, 4000},
            xticklabels={$D_L^{200}$,  $D_L^{1000}$,  $D_L^{2000}$, $D_L^{3000}$, $D_L^{4000}$},
            legend style={at={(axis cs:300,980)},anchor=south west}
                    ]

            \addplot[smooth,color=red,mark=x, thick] plot coordinates {
                (200,  50.086)
                (1000, 252.469)
                (2000, 602.828)
                (3000, 1058.498)
                (4000, 1632.383)

            };
            \addlegendentry{baseline}

            \addplot[smooth,mark=*,blue, thick]
                plot coordinates {
                  (200,   2.435)
                  (1000, 30.979)
                  (2000, 62.643)
                  (3000, 95.782)
                  (4000, 123.478)
                };
            \addlegendentry{MeTeoR}

         \end{axis}
    \end{tikzpicture}
%  }
 % \includegraphics[width=0.4\textwidth]{images/interval.png}
 \caption{Results of Experiment 2c
% Comparing reasoning time of our baseline and MeTeoR
% over datasets  $\D^m$ with $m=10,20,40$ and $60$.
 }
 \label{fig:intervals}
\end{figure}

The results of Experiment 2c are shown in Figure~\ref{fig:intervals}.
As anticipated,
the performance of the baseline  quickly degrades as the number
of temporal joins needed for SQL query evaluation increases,
whereas the performance of MeTeoR remains stable.
We see these
results as empirical validation for the need of specialised temporal
join algorithms, which are not provided off-the-shelf by SQL engines.

\section{Conclusion and Future Work}

We have presented the first practical algorithm and scalable implementation
for the full \MTL{} language. Our
system MeTeoR effectively combines optimised implementations of materialisation and
automata-based reasoning, and was able to successfully decide fact entailment over complex 
recursive programs
and large-scale datasets containing tens of millions of temporal facts.

We see many exciting avenues for future research. First, \citet{DBLP:conf/aaai/CucalaWGK21}
have recently extended \MTL{} with stratified negation as failure---a very useful feature
for applications---and provided an automata-based decision procedure. It would be interesting 
to extend MeTeoR to support stratified negation, which
 will require a non-trivial revision of our algorithm since materialisation within each
separate stratum may be non-terminating.
Second, it would be interesting to explore incremental reasoning techniques~\cite{walkega2019reasoning},
which are especially relevant to stream reasoning applications.
%  \citet{walega2021finitely} have recently proposed a class of \emph{finitely materialisable} \MTL{} programs for which 
% materialisation is guaranteed to terminate for every dataset. A natural extension of our
% work would be to extend our algorithm to first decide whether the input
% program is finitely materialisable, in which case we could simply run materialisation 
% until the input fact is entailed or a fixpoint is reached; this would identify many additional cases
% where automata reasoning is not required. 
Third, we aim to explore additional optimisations to further improve
scalability and also to apply MeTeoR in real-world applications  in collaboration with
our industrial partners.
\section*{Acknowledgments}
This work was supported by
the EPSRC projects OASIS (EP/S032347/1),
AnaLOG (EP/P025943/1),  and UK FIRES (EP/S019111/1),
the SIRIUS Centre for Scalable Data Access, and
Samsung Research UK.
\bibliographystyle{aaai}
\bibliography{references}

\newpage

\end{document}